\documentclass[10pt,twocolumn,letterpaper]{article}

\usepackage{cvpr}              %

\usepackage{graphicx}
\usepackage{amsmath}
\usepackage{amssymb}
\usepackage{booktabs}
\usepackage[accsupp]{axessibility}  %

\usepackage[pagebackref,breaklinks,colorlinks]{hyperref}

\usepackage[capitalize]{cleveref}
\crefname{section}{Sec.}{Secs.}
\Crefname{section}{Section}{Sections}
\Crefname{table}{Table}{Tables}
\crefname{table}{Tab.}{Tabs.}

\def\cvprPaperID{*****} %
\def\confName{CVPR}
\def\confYear{2023}

\usepackage{pifont}%
\usepackage{multirow}
\usepackage{color, colortbl}

\definecolor{Gray}{gray}{0.9}
\definecolor{LightCyan}{rgb}{0.88,1,1}

\usepackage[accsupp]{axessibility}

\usepackage[pagebackref,breaklinks,colorlinks]{hyperref}

\usepackage[capitalize]{cleveref}
\crefname{section}{Sec.}{Secs.}
\Crefname{section}{Section}{Sections}
\Crefname{table}{Table}{Tables}
\crefname{table}{Tab.}{Tabs.}

\usepackage{dblfloatfix}

\def\cvprPaperID{2173} %
\def\confName{CVPR}
\def\confYear{2023}

\begin{document}

\title{FREDOM: Fairness Domain Adaptation Approach to \\ Semantic Scene Understanding\vspace{-6mm}}

\author{
Thanh-Dat Truong$^{1}$, Ngan Le$^{1}$, Bhiksha Raj$^{2}$, Jackson Cothren$^{3}$, Khoa Luu$^{1}$\\
$^{1}$CVIU Lab, University of Arkansas, USA \quad
$^{2}$Carnegie Mellon University, USA  \\
$^{3}$Dep. of  Geosciences, University of Arkansas, USA \\
\tt\small \{tt032, thile, jcothre, khoaluu\}@uark.edu, bhiksha@cs.cmu.edu
\vspace{-4mm}
}

\maketitle

\begin{abstract}

Although Domain Adaptation in Semantic Scene Segmentation has shown impressive improvement in recent years, the fairness concerns in the domain adaptation have yet to be well defined and addressed. In addition, fairness is one of the most critical aspects when deploying the segmentation models into human-related real-world applications, e.g., autonomous driving, as any unfair predictions could influence human safety. In this paper, we propose a novel Fairness Domain Adaptation (FREDOM) approach to semantic scene segmentation. In particular, from the proposed formulated fairness objective, a new adaptation framework will be introduced based on the fair treatment of class distributions. Moreover, to generally model the context of structural dependency, a new conditional structural constraint is introduced to impose the consistency of predicted segmentation. Thanks to the proposed Conditional Structure Network, the self-attention mechanism has sufficiently modeled the structural information of segmentation. Through the ablation studies, the proposed method has shown the performance improvement of the segmentation models and promoted fairness in the model predictions. The experimental results on the two standard benchmarks, i.e., SYNTHIA $\to$ Cityscapes and GTA5 $\to$ Cityscapes, have shown that our method achieved State-of-the-Art (SOTA) performance\footnote{The implementation of FREDOM is available at \url{https://github.com/uark-cviu/FREDOM}}.

\end{abstract}

\vspace{-5mm}
\section{Introduction}

\begin{figure}[!ht]
    \centering
    \includegraphics[width=0.48\textwidth]{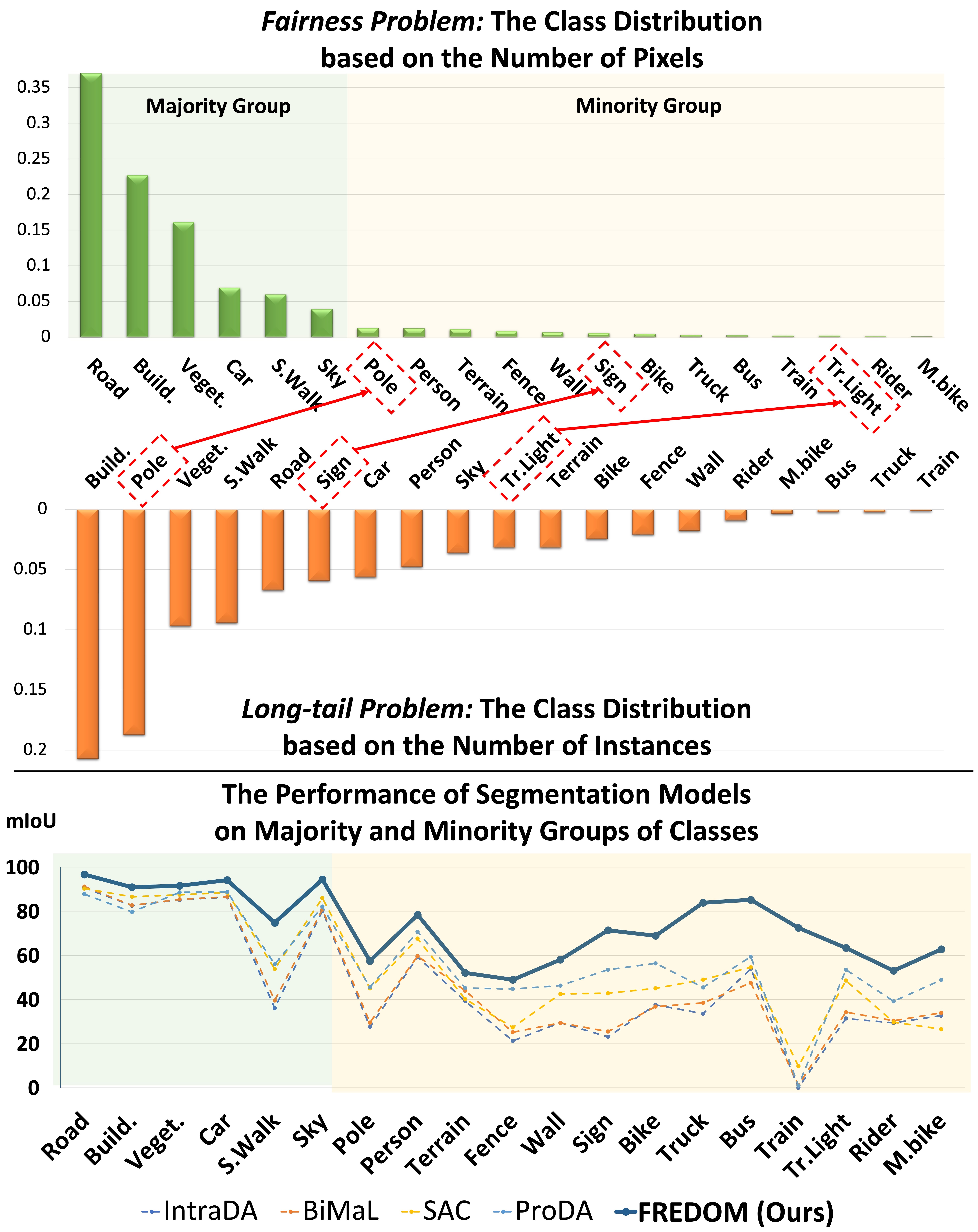}
    \vspace{-6mm}
    \caption{\textbf{The class distributions on Cityscapes are defined for Fairness problem and Long-tail problem.} In \textit{long-tail} problem, several head classes frequently exist in the dataset, e.g., Pole, Traffic Light, or Sign. Still, these classes belong to a minority group in the \textit{fairness} problem as their appearance on images does not occupy too many pixels. 
    Our FREDOM has promoted the fairness of models illustrated by an increase of mIoU on the minority group.}
    \vspace{-6mm}
    \label{fig:fair_long}
\end{figure}

Semantic segmentation has achieved remarkable results in a wide range of practical problems, including scene understanding, autonomous driving, and medical imaging, by using deep learning models, e.g., Convolutional Neural Networks (CNN) \cite{chen2018deeplab, chen2018encoder, lin2017refinenet}, Transformers \cite{xie2021segformer}. Despite the phenomenal achievement, these data-driven approaches still need to improve in treating the prediction of each class. In particular, the segmentation models typically treat unfairly between classes in the dataset according to the class distributions. It is known as the fairness problem of semantic segmentation. The unfair predictions of segmentation models can lead to severe problems, e.g., in autonomous driving, unfair predictions may result in wrong decisions in motion planning control and therefore affect human safety. Moreover, the fairness issue of segmentation models is even well observed or exaggerated when the trained models are deployed into new domains. Many prior works alleviate the performance drop on new domains by using unsupervised domain adaptation, but these approaches do not guarantee the fairness property.

\begin{figure}[!t]
    \centering
    \includegraphics[width=0.48\textwidth]{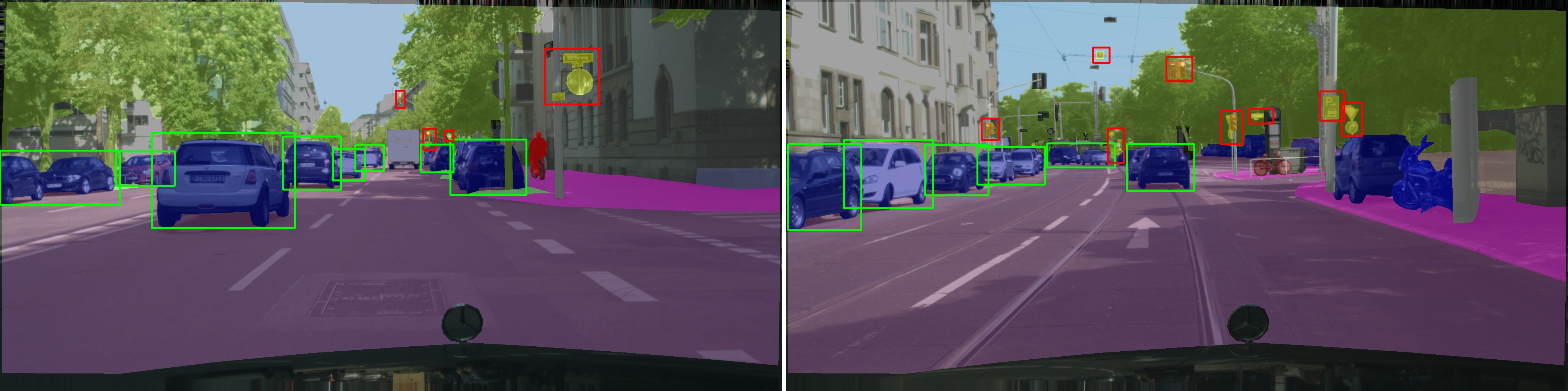}
    \vspace{-7mm}
    \caption{\textbf{Illustration of the Presence of Classes between Major (green boxes) and Minor (red boxes) Groups}. Classes in the minority group typically occupy fewer pixels than the ones in the majority group (Best view in color and $2\times$ zoom).}
    \label{fig:major_minor_fig}
    \vspace{-6mm}
\end{figure}

There needs to be more attention on addressing the fairness issue in semantic segmentation under the supervised or domain adaptation settings. Besides, the definition of fairness in semantic segmentation needs to be better defined and, therefore, often needs clarification with the long-tail issue in segmentation. In particular, the \textbf{\textit{long-tail problem}} in segmentation is typically caused by \textbf{\textit{the number of existing instances}} of each class in the dataset \cite{wang2021seesaw, DBLP:conf/aaai/Ting21}. Meanwhile, the \textbf{\textit{fairness problem}} in segmentation is considered for \textbf{\textit{the number of pixels}} of each class in the dataset.
Although there could be a correlation between fairness and long-tail problems, these two issues are distinct. For example, several objects constantly exist in the dataset, but their presence often occupies only tiny regions of the given image (containing a small number of pixels), 
e.g., the Pole, which is a head class in Cityscapes, accounts for over  $20\%$ of instances while the number of pixels does only less than $0.01\%$ of pixels. 
Hence, upon the fairness definition, it should belong to the minor group of classes as its presence does not occupy many pixels in the image. 
Another example is Person, which accounts for over $5\%$ of instances, while the number of pixels does only less than $0.01\%$ of pixels. 
Traffic Lights or Signs also suffer a similar problem.
Fig. \ref{fig:major_minor_fig} illustrates the appearance of classes in the majority and minority groups.
Therefore, although instances of these classes constantly exist in the dataset, these are still being mistreated by the segmentation model. Fig. \ref{fig:fair_long} illustrates the class distributions defined based on long-tail and fairness, respectively.

Several works reduce the class imbalance effects using weighted (balanced) cross entropy \cite{Cui_2019_CVPR, wang2021seesaw, DBLP:conf/aaai/Ting21}, focal loss \cite{Araslanov:2021:DASAC}, data augmentation or rare-class sampling techniques \cite{Araslanov:2021:DASAC, daformer}. Still, these need to address the fairness problem directly. Indeed, many prior domain adaptation methods  \cite{hoffman2016fcns, tsai2018learning, tsai2019domain, chen2019domain, 9956335, 9108692, 9897440, truong2022conda} have been used to improve the overall performance. However, these methods often ignore unfair effects produced by the model caused by the imbalanced class distribution. Besides, in some adaptation approaches using entropy minimization \cite{vu2019advent, pan2020unsupervised}, the model's bias caused by the class imbalance between majority and minority groups is even exaggerated \cite{truong2021bimal, Chen_2019_ICCV}.
Meanwhile, other approaches using re-weighted or focal loss \cite{Araslanov:2021:DASAC} often assume pixel independence and then penalize the loss contribution of each pixel individually and ignore the structural information of images. Then, pixel independence is relaxed by adopting the Markovian assumption \cite{Zheng_2015_ICCV, chen2018deeplab} to model segmentation structures based on neighbor pixels. 
In the \textit{scope of our work}, we are interested in addressing the fairness problem in semantic segmentation between classes under the unsupervised domain adaptation setting. It should be noted that our interested problem is practical. In real-world applications (e.g., autonomous driving), deep learning models are typically deployed into new domains compared to the training dataset. Then, unsupervised domain adaptation plays a role in bridging the gap between the two domains.

\noindent
\textbf{Contributions of This Work:} This work presents a novel Unsupervised \textbf{F}ai\textbf{r}n\textbf{e}ss \textbf{Dom}ain Adaptation (FREDOM) approach to semantic segmentation. To the best of our knowledge, this is one of the first works to address the fairness problem in semantic segmentation under the domain adaptation setting. Our contributions can be summarized as follows. First, the new fairness objective is formulated for semantic scene segmentation. Then, based on the fairness metric, we propose a novel fairness domain adaptation approach based on the fair treatment of class distributions. Second, the novel Conditional Structural Constraint is proposed to model the structural consistency of segmentation maps. 
Thanks to our introduced Conditional Structure Network, the spatial relationship and structure information are well modeled by the self-attention mechanism. Significantly, our structural constraint relaxes the assumption of pixel independence held by prior approaches and generalizes the Markovian assumption by considering the structural correlations between all pixels. Finally, our ablation studies have shown the effectiveness of different aspects in our approach to the fairness improvement of segmentation models. Through experiments, our FREDOM has promoted the fairness property of segmentation models and achieved state-of-the-art (SOTA) performance on two standard benchmarks of unsupervised domain adaptation, i.e., SYNTHIA $\to$ Cityscapes and GTA5 $\to$ Cityscapes. 

\vspace{-1mm}
\section{Related Work}
\vspace{-1mm}

Unsupervised Domain Adaptation (UDA) in Semantic Segmentation is a vital research topic as its ability to reduce the necessity for massive volumes of labeled data. 
Adversarial learning \cite{chen2018road, hong2018CVPR, tsai2018learning, ganin2015unsupervised, long2015learning, tzeng2017adversarial}, and self-supervised training \cite{Araslanov:2021:DASAC, zhang2021prototypical, daformer, procst} are common approaches to UDA.

\noindent \textbf{Adversarial Learning} is a common approach to UDA in semantic segmentation. The model is simultaneously trained on source and target domains in this approach. Hoffman \etal \cite{hoffman2016fcns} introduced the first adversarial approach to UDA in segmentation. Then, Chen \etal \cite{chen2017no} improved the model by utilizing pseudo labels in parallel with the global and class-wise adaptation learning process. The distillation loss with spatial-aware model \cite{chen2018road} proposed by Chen \etal has been utilized to improve the spatial structures of segmentation. Other methods have approached the UDA problem by using image translation \cite{zhu2017unpaired, murez2018CVPR, hoffman18a}. SPIGAN \cite{lee2018spigan} embed depth information as its privileged information to improve the UDA model for semantic segmentation. Similarly, Vu \etal \cite{vu2019dada} proposed a depth-aware framework using privileged depth information. Vu \etal \cite{vu2019advent} presented the first adversarial entropy minimization approach to UDA in segmentation. Then, \cite{pan2020unsupervised, 10.1145/3474085.3475174} presented a curriculum adaptation training from easy to complex samples ranked by the entropy level. Truong \etal \cite{truong2021bimal, 9874820} improved the performance of segmentation models by introducing a bijective maximum likelihood approach. 

\noindent \textbf{Self-supervised Approach} has gained a SOTA performance in UDA in semantic segmentation in recent years \cite{zou2018unsupervised, Araslanov:2021:DASAC, zhang2021prototypical, daformer, procst}. In self-training approaches, a new model is trained on unlabeled data using pseudo-labels derived from a trained model. Araslanov \etal \cite{Araslanov:2021:DASAC} proposed an augmentation consistency approach to automatically evolve pseudo labels without using further training rounds. Zhang \etal \cite{zhang2021prototypical} introduced a knowledge distillation approach to improving the performance of models while also correcting the soft pseudo labels online. Hoyer \etal \cite{daformer} improved the performance of UDA via a new Transformer-based backbone and training recipe. Then, \cite{daformer} is further improved by introducing a context-aware high-resolution framework that utilizes the advantages of small high-resolution crops for maintaining precise segmentation and large low-resolution crops for capturing context dependencies \cite{hoyer2022hrda}.

\noindent \textbf{Class Imbalance Approaches:} 
Jiawei \etal \cite{ren2020balanced} presented a balanced Softmax loss that helps reduce labels' distribution shift and alleviates the long-tail issue. 
Wang \etal \cite{wang2021seesaw} proposed a Seesaw loss that reweights the contributions of gradients produced by positive and negative instances of a class by using two regularizers, i.e., mitigation and compensation. Ziwei \etal \cite{liu2019largescale} proposed an algorithm that handles imbalanced classification, few-shot learning, and open-set recognition using dynamic meta-embedding. Chu \etal \cite{chu2021learning} proposed a stochastic training scheme for semantic segmentation, which improves the learning of debiased and disentangled representations. Szabo \etal \cite{szabo2021tilted} proposed tilted cross-entropy loss to reduce the performance differences, which promotes fairness among the target classes.

\vspace{-2mm}
\section{The Proposed Fairness Domain Adaptation Approach to Semantic Segmentation}
\vspace{-1mm}

Let $\mathbf{x}_s \in \mathcal{X}_s$ and $\mathbf{\hat{y}}_s \in \mathcal{Y}_s$ be an input image and its corresponding segmentation label in the source domain drawn from the source distribution $p_s$,
$\mathbf{x}_t \in \mathcal{X}_t$ and $\mathbf{\hat{y}}_t \in \mathcal{Y}_t$ be the input image and the segmentation label in the target domain drawn from the target distribution $p_t$. 
In unsupervised domain adaptation, %
the ground-truth segmentation $\mathbf{\hat{y}}_t$ of image $\mathbf{x}_t$ is not available.
Let $F: \mathcal{X} = \mathcal{X}_s \cup \mathcal{X}_t \to \mathcal{Y} = \mathcal{Y}_s \cup \mathcal{Y}_t$ be the deep network parameterized by $\theta$ that maps the input image $\mathbf{x} \in \mathcal{X}$ into the segmentation $\mathbf{y} \in \mathcal{Y}$, i.e $\mathbf{y}_s = F(\mathbf{x}_s, \theta)$, and $\mathbf{y}_t = F(\mathbf{x}_t, \theta)$.
The standard domain adaptation can be mathematically formulated as in Eqn. \eqref{eqn:original_optimization}.
\vspace{-4.5mm}
\begin{equation} 
\label{eqn:original_optimization}
\footnotesize
    \theta^* = \arg\min_{\theta} \left[\mathbb{E}_{\mathbf{x}_s, \mathbf{\hat{y}}_s \sim p_s(\mathbf{y}_s, \mathbf{\hat{y}}_s}) \mathcal{L}_s(\mathbf{y}_s, \mathbf{\hat{y}}_s)  + \mathbb{E}_{\mathbf{x}_t \sim p_t(\mathbf{x}_t)} \mathcal{L}_t(\mathbf{y}_t)\right] 
\end{equation}
where $\mathcal{L}_s$ is the supervised cross-entropy (CE) loss in the source domain.
Meanwhile, $\mathcal{L}_t$ is the unsupervised learning loss in the target domain that can be defined as the adversarial loss \cite{tsai2018learning, tsai2019domain, vu2019advent, pan2020unsupervised}, 
or the self-supervised loss %
\cite{Araslanov:2021:DASAC, zhang2021prototypical, daformer}. In recent studies, the self-supervised loss defined by the cross-entropy loss with pseudo labels has achieved SOTA performance and outperformed other prior methods. Therefore, our proposed approach also defines $\mathcal{L}_t$ as the self-supervised loss \cite{Araslanov:2021:DASAC, daformer} with the novel fairness guarantee.

\subsection{The Fairness Objective Function}

Under the fairness constraint in semantic segmentation, the performance of each class should be equally treated by the deep model. Thus, the goal of fairness in semantic segmentation can be defined as in Eqn. \eqref{eqn:fairness}.
\begin{equation} \label{eqn:fairness}
\footnotesize
\begin{split}
    \arg\min_{\theta} \sum_{c_i, c_j}\Big|\mathbb{E}_{\mathbf{x} \in \mathcal{X}}\sum_{k}\mathcal{L}(y^k = c_i) - \mathbb{E}_{\mathbf{x} \in \mathcal{X}}\sum_{k}\mathcal{L}(y^k = c_j)\Big|
\end{split}
\end{equation}
where $y^k$ denotes the $k^{th}$ pixel of the segmentation $\mathbf{y}$, 
$c_i$ and $c_j$ are the class categories, i.e, $c_i, c_j \in [1..C]$ (where $C$ is the number of classes),
$\mathcal{L}$ is the loss function measuring the error rates of predictions. 
Formally, for all pairs of classes in the dataset,  Eqn. \eqref{eqn:fairness} aims to minimize the difference in the error rates produced by the model between classes. Therefore, it guarantees all classes in the dataset are treated equally. Eqn. \eqref{eqn:fairness} can be further derived as in Eqn. \eqref{eqn:fairness_less_adapt}.
\begin{equation} \label{eqn:fairness_less_adapt}
\footnotesize
\begin{split}
&\sum_{c_i, c_j}\Big|\mathbb{E}_{\mathbf{x} \in \mathcal{X}}\sum_{k}\mathcal{L}(y^k_s = c_i) - \mathbb{E}_{\mathbf{x} \in \mathcal{X}}\sum_{k}\mathcal{L}(y^k_s = c_j)\Big|\\
&\leq \sum_{c_i, c_j}\Big(\mathbb{E}_{\mathbf{x} \in \mathcal{X}}\sum_{k}\mathcal{L}(y^k_s = c_i) + \mathbb{E}_{\mathbf{x} \in \mathcal{X}}\sum_{k}\mathcal{L}(y^k_s = c_j)\Big)\\
&= 2C\mathbb{E}_{\mathbf{x} \in \mathcal{X}}\mathcal{L}(\mathbf{y}) 
= 2C\Big[\mathbb{E}_{\mathbf{x}_s \in \mathcal{X}_s} \mathcal{L}_s(\mathbf{y}_s)  + \mathbb{E}_{\mathbf{x}_t \in \mathcal{X}_t} \mathcal{L}_t(\mathbf{y}_t)\Big] \\
&=2C\Big[\mathbb{E}_{\mathbf{x}_s, \mathbf{\hat{y}_s} \sim p_s(\mathbf{y}_s, \mathbf{\hat{y}}_s}) \mathcal{L}_s(\mathbf{y}_s, \mathbf{\hat{y}}_s)  + \mathbb{E}_{\mathbf{x}_t \sim p_t(\mathbf{x}_t)} \mathcal{L}_t(\mathbf{y}_t)\Big] 
\end{split}
\end{equation}
From Eqn. \eqref{eqn:fairness_less_adapt}, we can observe that the fairness objective in Eqn. \eqref{eqn:fairness} is bounded by the standard optimization of domain adaptation in Eqn. \eqref{eqn:original_optimization}.
Although optimizing the standard domain adaptation as in Eqn.  \eqref{eqn:original_optimization} could impose the constraint of fairness under the upper bound in Eqn. \eqref{eqn:fairness_less_adapt}, the imbalance class distributions of pixels cause the model to behave unfairly between classes when optimizing Eqn. \eqref{eqn:original_optimization}. In particular,
Eqn. \eqref{eqn:original_optimization} can be rewritten as follows,
\begin{equation} \label{eqn:original_optimization_rewrite}
\scriptsize
    \begin{split}
    &\arg\min_{\theta} \Bigg[\int \mathcal{L}_s(\mathbf{y}_s, \mathbf{\hat{y}}_s)p_s(\mathbf{y}_s) p_s(\mathbf{\hat{y}}_s)d\mathbf{y}_s d\mathbf{\hat{y}}_s 
    + \int \mathcal{L}_t(\mathbf{y}_t)p_t(\mathbf{y}_t)d\mathbf{y}_t\Bigg]\\
    &=\arg\min_{\theta} \Bigg[\int \sum_{k=1}^N\mathcal{L}_s(y^k_s, \hat{y}^k_s) p_s(y^k_s)p_s(\mathbf{y}^{\setminus k}_s | y^k_s)p_s(\mathbf{\hat{y}}_s)d\mathbf{y}_s d\mathbf{\hat{y}}_s  \\
    &\qquad\qquad\qquad\qquad\qquad\qquad\qquad  
    + \int \sum_{k=1}^N\mathcal{L}_t(y^k_t) p_t(y^k_t)p_t(\mathbf{y}^{\setminus k}_t | y^k_t)d\mathbf{y}_t\Bigg]
    \end{split}
\end{equation}
where $N$ is the total number of pixels in the image, 
$y^k_s$ and  $y^k_t$ are the $k^{th}$ pixel of predicted segmentations in source and target domains,
$\mathbf{y}^{\setminus k}_s$ and $\mathbf{y}^{\setminus k}_t$ are predicted segmentations without the $k^{th}$ pixel in source and target domains, $p_s(y^k)$ and $p_t(y^k)$ are the class distributions of pixels in the source and target domains.
The class distributions are computed based on the number of pixels of each class in the dataset.
The terms $p_s(\mathbf{y}^{\setminus k}_s | y^k_s)$ and $p_t(\mathbf{y}^{\setminus k}_t | y^k_t)$ are conditional structure constraints of $\mathbf{y}_s^{\setminus k}$ and $\mathbf{y}_t^{\setminus k}$ on $y^k_s$ and $y^k_t$.

\noindent
\textbf{From imbalance distributions to unfair predictions:}
In practice, the class distributions of pixels $p_s(y_s^k)$ and $p_t(y_t^k)$ suffer imbalance problems as shown in Fig. \ref{fig:fair_long}.
When the model is learned by the gradient descent method, the model behaves inequitably between classes. In particular, let us consider the behavior of gradients produced by the gradient descent learning method. Formally, let $c_i$ and $c_j$ be the two classes in the dataset and $p_s(y_s^k=c_i) << p_s(y^k_s=c_j)$. The gradients produced for each class with respect to the predictions can be formed as in Eqn. \eqref{eqn:inequality}.
\begin{equation} \label{eqn:inequality}
\scriptsize
\begin{split}
    &\left|\left|\frac{\partial \int \sum_{k=1}^{N}\mathcal{L}_s(y^k_s, \hat{y}^k_s) p_s(y^k_s=c_i)p_s(\mathbf{y}^{\setminus k}_s | y^k_s )p_s(\mathbf{\hat{y}}_s)d\mathbf{y}_s d\mathbf{\hat{y}}_s}{\partial \mathbf{y}^{(c_i)}_{s}}\right|\right| \ll \\
    &\quad\quad \left|\left|\frac{\partial \int \sum_{k=1}^{N}\mathcal{L}_s(y^k_s, \hat{y}^k_s) p_s(y^k_s=c_j)q_s(\mathbf{y}^{\setminus k}_s | y^k_s )p_s(\mathbf{\hat{y}}_k)d\mathbf{y}_s d\mathbf{\hat{y}}_s}{\partial \mathbf{y}^{(c_j)}_{s}}\right|\right|
\end{split}
\end{equation}
where $||.||$ is the magnitude of the vector,
$\mathbf{y}^{(c_i)}_s$ and $\mathbf{y}^{(c_j)}_s$ represent the predicted probabilities of label $c_i$ and $c_j$, respectively.
As shown in Eqn. \eqref{eqn:inequality}, 
the model inclines to produce significant gradient updates of the classes having a large population in the distributions (\textit{a majority group}); meanwhile, the gradient updates of the class having a small population in the distributions (\textit{a minority group}) are minor and dominated by the gradients of majority groups. 
Similar behavior can also be observed in the target domain.

\subsection{The Proposed Fairness Adaptation Approach}

\begin{figure*}
    \centering
    \includegraphics[width=0.75\textwidth]{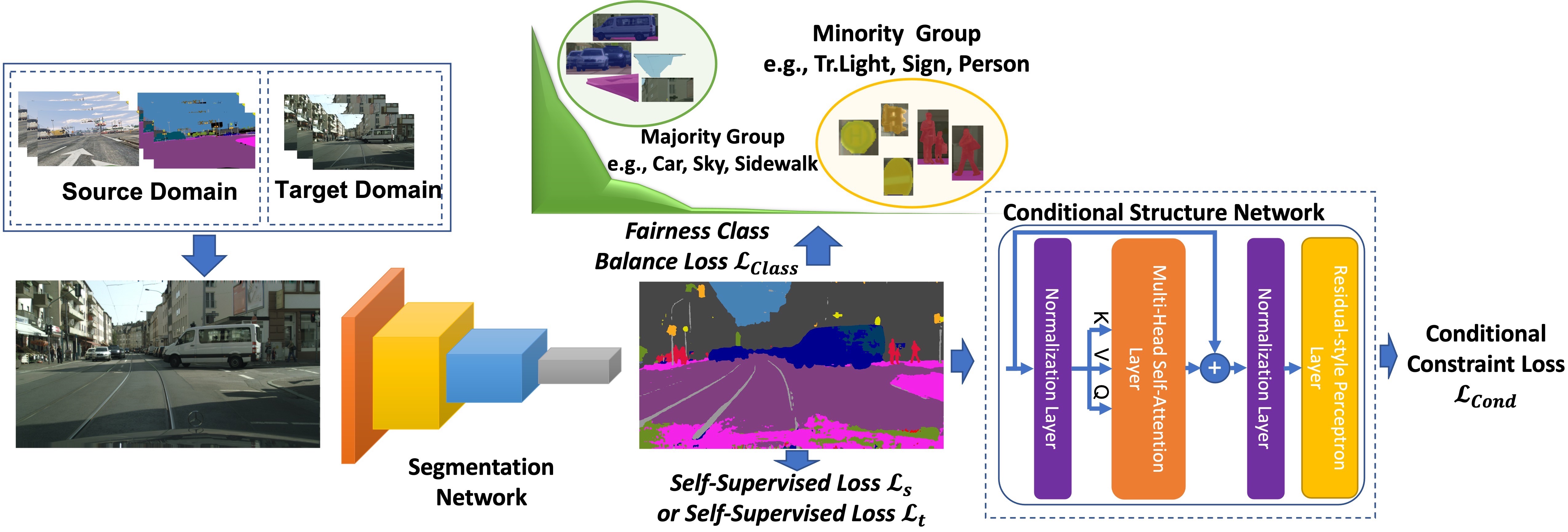}
    \vspace{-3mm}
    \caption{\textbf{The Proposed Fairness Framework.}
    The predictions of the inputs sampled from the source or target domains are penalized by the supervised loss $\mathcal{L}_s$ or the self-supervised loss $\mathcal{L}_t$, respectively. Then, the predictions are imposed by the fairness class balance loss $\mathcal{L}_{Class}$ followed by the Conditional Constraint Loss $\mathcal{L}_{Cond}$ computed via a Conditional Structure Network (Best view in color).
    }
    \vspace{-6mm}
    \label{fig:fair_da_framework}
\end{figure*}

As discussed in the previous section, the fairness problem 
is typically caused by imbalanced class distributions. 
Therefore, to address the fairness problem, we first assume that there exists an ideal distribution $p'_s(\mathbf{y}_s)$ and $p'_t(\mathbf{y}_t)$ so that the model trained on the ideal data distributions behave fairly between classes. 
It should be noted that we assume the ideal data distribution to frame and navigate our proposed approach to the fairness domain adaptation in semantic segmentation.
Then, the ideal data distributions will be relaxed later and there is no requirement for the ideal data distribution during the training process.
Formally, learning the adaptation framework of Eqn. \eqref{eqn:original_optimization} under the ideal data distribution can be formulated as in Eqn. \eqref{eqn:opt_from_ideal}.
\begin{equation} \label{eqn:opt_from_ideal}
\footnotesize
\begin{split}
      &\arg\min_{\theta} \Bigg[\mathbb{E}_{\mathbf{x}_s \sim p_s(\mathbf{y}_s), \mathbf{\hat{y}_s} \sim p_s(\mathbf{\hat{y}}_s)} \mathcal{L}_s(\mathbf{y}_s, \mathbf{\hat{y}}_s)\frac{p'_s(\mathbf{y}_s) p'_s(\mathbf{\hat{y}}_s)}{p_s(\mathbf{y}_s)p_s( \mathbf{\hat{y}}_s)}  \\
    &\qquad\qquad\qquad\qquad\qquad\qquad
    + \mathbb{E}_{\mathbf{x}_t \sim p_t(\mathbf{x}_t)} \mathcal{L}_t(\mathbf{y}_t)\frac{p'_t(\mathbf{y}_t)}{p_t(\mathbf{y}_t)}\Bigg] 
\end{split}
\end{equation}
The fraction between ideal and real data distributions, i.e. $\frac{p'_s(\mathbf{y}_s) p'_s(\mathbf{\hat{y}}_s)}{p_s(\mathbf{y}_s)p_s( \mathbf{\hat{y}}_s)}$ and $\frac{p'_t(\mathbf{y}_t)}{p_t(\mathbf{y}_t)}$, can be interpreted as the complement of the model needed to be improved to achieve fairness against the imbalanced data.
It should be noted that $p'_s(\mathbf{\hat{y}}_s)$ and $p_s( \mathbf{\hat{y}}_s)$ are constants as they are distributed over segmentation labels, 
so these could be excluded during training. %
Then, Eqn. \eqref{eqn:opt_from_ideal} can be further derived as follows,
\begin{equation} \label{eqn:opt_from_ideal_sum}
\footnotesize
\begin{split}
    & \arg\min_{\theta} \Bigg[\mathbb{E}_{\mathbf{x}_s \sim p_s(\mathbf{y}_s), \mathbf{\hat{y}_s} \sim p_s(\mathbf{\hat{y}}_s)} \sum_{k=1}^{N} \mathcal{L}_s(y^k_s, \hat{y}^k_s) \frac{p'_s(y^k_s)p'_s(\mathbf{y}^{\setminus k}_s|y^k_s)}{p_s(y^k_s)p_s(\mathbf{y}^{\setminus k}_s |y^k_s)}  \\
    &\qquad\qquad\qquad\qquad + \mathbb{E}_{\mathbf{x}_t \sim p_t(\mathbf{x}_t)} \sum_{k=1}^{N}\mathcal{L}_t(y^k_t) \frac{p'_t(y^k_t)p'_t(\mathbf{y}^{\setminus k}_t|y^k_t)}{p_t(y^k_t)p_t(\mathbf{y}^{\setminus k}_t |y^k_t)} \Bigg]
\end{split}
\end{equation}
As shown in Eqn. \eqref{eqn:opt_from_ideal_sum}, if the conditional structure fractions $\frac{p'_s(\mathbf{y}^{\setminus k}_s|y^k_s)}{p_s(\mathbf{y}^{\setminus k}_s |y^k_s)}$ and $\frac{p'_t(\mathbf{y}^{\setminus k}_t|y^k_t)}{p_t(\mathbf{y}^{\setminus k}_t |y^k_t)}$ are ignored, Eqn. \eqref{eqn:opt_from_ideal_sum} becomes a special case of the weighted class balanced loss \cite{Cui_2019_CVPR, wang2021seesaw}. %
However, conditional structure plays a vital role in semantic segmentation as it provides the constraints and correlation of structures among objects in images. The ignorance of conditional structure fractions could lower the
performance of segmentation models.
In addition, although the input images of the source and target domains can vary significantly in appearance due to the distribution shift, their segmentation maps between two domains share similar class distributions and structural information \cite{tsai2018learning, tsai2019domain, truong2021bimal}. Hence, the distribution of segmentation in the target domain $p_t(\cdot)$ can be practically approximated by distribution in the source domain, i.e., $\frac{p'_t(\mathbf{y}_t)}{p_t(\mathbf{y}_t)} = \frac{p'_s(\mathbf{y}_t)}{p_s(\mathbf{y}_t)}$.
In summary, by taking the log of Eqn. \eqref{eqn:opt_from_ideal_sum}, the learning process can be formed as follows 
(the derivation of Eqn. \eqref{eqn:take_log} is detailed in the supplementary):
\vspace{-6mm}
\begin{equation} \label{eqn:take_log}
    \scriptsize
    \begin{split}
        &\theta^* \simeq \arg\min_{\theta} \Bigg[\mathbb{E}_{\mathbf{x}_s \sim p_s(\mathbf{x}_s), \mathbf{\hat{y}_s} \sim p_s(\mathbf{\hat{y}}_s)} \mathcal{L}_s(\mathbf{y}_s, \mathbf{\hat{y}}_s) + \mathbb{E}_{\mathbf{x}_t \sim p_t(\mathbf{x}_t)} \mathcal{L}_t(\mathbf{y}_t)\\
        &+\frac{1}{N}\sum_{k=1}^{N}
        \Bigg(\mathbb{E}_{\mathbf{x}_s \sim p_s(\mathbf{x}_s)} \log\left(\frac{p'_s(y^k_s)}{p_s(y^k_s)}\right) 
        +\mathbb{E}_{\mathbf{x}_t \sim p_t(\mathbf{x}_t)} \log\left(\frac{p'_s(y^k_t)}{p_s(y^k_t)}\right)\\
        &+\mathbb{E}_{\mathbf{x}_s \sim p_s(\mathbf{x}_s)} \log\left(\frac{p'_s(\mathbf{y}^{\setminus k}_s |y^k_s)}{p_s(\mathbf{y}^{\setminus k}_s |y^k_s)}\right)
        +\mathbb{E}_{\mathbf{x}_t \sim p_t(\mathbf{x}_t)} \log\left(\frac{p'_s(\mathbf{y}^{\setminus k}_t|y^k_t)}{p_s(\mathbf{y}^{\setminus k}_t |y^k_t)}\right)\Bigg)\Bigg]
        \vspace{-2mm}
    \end{split}
\end{equation}

In summary, there are three terms in the learning objective of our FREDOM approach. 
Hence, several properties are brought into the learning process that can be observed.

\noindent
\textbf{Domain Adaptation Objective} The first two terms stand for the objective of domain adaptation. While $\mathcal{L}_s$ learns to a segment on the source domain in the supervised fashion, $\mathcal{L}_t$ aims to unsupervised adapt knowledge to the target domain.

\noindent
\textbf{Fairness Treatment from Class Distributions} The next two terms, i.e, $\log\left(\frac{p'_s(y^k_t)}{p_s(y^k_t)}\right)$ and $\log\left(\frac{p'_s(y^k_t)}{p_s(y^k_t)}\right)$, denoted as the $\mathcal{L}_{Class}$, impose the behavior of the model with respect to the class distribution. In particular, these constraints aim to regularize the predictions of classes so that the model should behave fairly between classes with respect to the class distribution. Under the ideal data distribution assumption, the model is expected to equally treat predictions of all classes. Thus, to achieve the desired goal, the distributions of pixel classes should be uniformly distributed. Therefore, we adopt the uniform distribution of the class distribution $p'_s(y^k_s)$, i.e., $p'_s(y^k_s) = \frac{1}{C}$ where C is the number of classes.

\noindent
\textbf{Conditional Structure Constraint}
The last two terms, i.e., $\log\left(\frac{p'_s(\mathbf{y}^{\setminus k}_s |y^k_s)}{p_s(\mathbf{y}^{\setminus k}_s |y^k_s)}\right)$ and $\log\left(\frac{p'_s(\mathbf{y}^{\setminus k}_t|y^k_t)}{p_s(\mathbf{y}^{\setminus k}_t |y^k_t)}\right)$, denoted as $\mathcal{L}_{Cond}$, impose the conditional structure of the predicted semantic segmentation.
This condition plays a role as a metric to measure the structural consistency of predicted segmentation maps with respect to the one under the ideal distributions where the model behaves fairly.
Modeling the conditional structure, i.e., $p_s(\mathbf{y}^{\setminus k}_s |y^k_s)$, is a challenging problem. Several prior works 
modeled structural constraints by adopting the Markovian assumption \cite{Zheng_2015_ICCV, chen2018deeplab}
where the models only 
consider the correlation between %
the current pixel with its neighbor pixels.
However, the smoothness of predicted segmentation maps is highly dependent on the window size used in Markovian approaches (the number of neighbor pixels being selected). 
In our work, to sufficiently capture the conditional structural constraint, instead of modeling only neighborhood dependencies as Markovian approaches, we generalize it by modeling $p_s(\mathbf{y}^{\setminus k}_s |y^k_s)$ via a conditional structure network (detailed in Sec. \ref{sec:Cond_Network}) 
to consider the correlation between all pixels in the segmentation. 

\noindent
\textbf{Relaxation of Ideal Data Distribution}
One of the key challenging problems in optimizing Eqn. \eqref{eqn:take_log} is that the conditional ideal data distributions $p'_s(\mathbf{y}^{\setminus k}_s |y^k_s)$ and $p'_s(\mathbf{y}^{\setminus k}_t|y^k_t)$ are not available.
Therefore, instead of directly optimizing these terms, let us consider the tight bound as in Eqn. \eqref{eqn:bound}.
\begin{equation}
\label{eqn:bound}
\footnotesize
\begin{split}
    &\mathbb{E}_{\mathbf{x}_s \sim p_s(\mathbf{x}_s)} \log\left(\frac{p'_s(\mathbf{y}^{\setminus k}_s |y^k_s)}{p_s(\mathbf{y}^{\setminus k}_s |y^k_s)}\right) 
    + \mathbb{E}_{\mathbf{x}_t \sim p_t(\mathbf{x}_t)} \log\left(\frac{p'_s(\mathbf{y}^{\setminus k}_t|y^k_t)}{p_s(\mathbf{y}^{\setminus k}_t |y^k_t)}\right)\\
    &\leq -\Big[\mathbb{E}_{\mathbf{x}_s \sim p_s(\mathbf{x}_s)} \log p_s(\mathbf{y}^{\setminus k}_s |y^k_s) 
    + \mathbb{E}_{\mathbf{x}_t \sim p_t(\mathbf{x}_t)} \log p_s(\mathbf{y}^{\setminus k}_t |y^k_t)\Big]
\end{split}
\end{equation}
With any form of ideal distribution $p'_s(\cdot)$, Eqn. \eqref{eqn:bound} always hold due to $\log p'_s(\cdot) \leq 0$. Hence, optimizing Eqn. \eqref{eqn:bound} also ensure the conditional structural constraint in Eqn. \eqref{eqn:take_log} imposed due to the upper bound of Eqn. \eqref{eqn:bound}. Therefore, the demand for ideal data distribution is relaxed.
Fig. \ref{fig:fair_da_framework} illustrates our proposed fairness domain adaptation framework.

\section{The Conditional Structure Network} \label{sec:Cond_Network}

The conditional structural constraint $p_s(\mathbf{y}^{\setminus k}_s |y^k_s)$ can be learned on the source dataset due to the availability of the ground-truth segmentation in the source domain.
Formally, let $p_s(\mathbf{y}^{\setminus k}_s |y^k_s)$ be modeled by the conditional structure network $G$ with parameters $\Theta$. Then the conditional structure network can be auto-regressively formed as follows:
\vspace{-1mm}
\begin{equation} \label{eqn:auto_erg_rnn}
\footnotesize
\begin{split}
    &\arg\min_{\Theta} \mathbb{E}_{\mathbf{y}_s \in \mathcal{Y}_s} -\log p_s(\mathbf{y}^{\setminus k}_s |y^k_s, \Theta) \\
    =&\arg\min_{\Theta} \mathbb{E}_{\mathbf{y}_s \in \mathcal{Y}_s} \sum_{i=1}^{N-1} -\log p_s(y^{\sigma^k_i} | y^{\sigma^k_{i-1}}, ..., y^{\sigma^k_1}, y^k_s, \Theta)
    \vspace{-3mm}
\end{split}
\end{equation}
where $\sigma^k$ is the permutation of $\{1...N\} \setminus \{k\}$. Eqn. \eqref{eqn:auto_erg_rnn} could be modeled by Recurrent Neural Networks \cite{oord2016pixel}. However, directly adopting recurrent approaches remains some potential limitations.
Particularly, as the recurrent approaches use a pre-defined permutation of regressive orders, it requires different conditional structure models for different initial pixel conditions, e.g., $p_s(\mathbf{y}^{\setminus k_1}_s |y^{k_1}_s)$ and $p_s(\mathbf{y}^{\setminus k_2}_s |y^{k_2}_s)$ should be modeled two different models.
This problem could be alleviated by considering the permutation of regressive order
as an network's input. %
However, learning a single network to model conditional structural constraints of different permutations is a heavy task and ineffective.

Instead of regressively forming $p_s(\mathbf{y}^{\setminus k}_s |y^k_s)$, we propose to model $p_s(\mathbf{y}^{\setminus k}_s |y^k_s)$ in the parallel fashion. 
Particularly, let $\mathbf{m}$ be the binary masked matrix of $\mathbf{y}_s$, where
the values of one and zero indicate a given pixel (unmasked pixel) and an unknown pixel (masked pixel), respectively. 
Then, the conditional structure $p_s(\mathbf{y}^{\setminus k}_s |y^k_s)$ can be rewritten as $p_s(\mathbf{y}_s \odot (\mathbf{1} - \mathbf{m}) | \mathbf{y}_s \odot \mathbf{m})$, 
where $\odot$ is the element-wise product and the mask $\mathbf{m}$ contains only one unmasked pixel, i.e., the given $k^{th}$ pixel ($m^k = 1$).
Learning the conditional structure constraint via binary mask $\mathbf{m}$ can be formed as: 
\begin{equation} \label{eqn:bert_gpt}
\footnotesize
\begin{split}
    \arg\min_{\Theta} \mathbb{E}_{\mathbf{y}_s \in \mathcal{Y}_s, \mathbf{m} \in \mathcal{M}} -\log p_s(\mathbf{y}_s \odot (\mathbf{1} - \mathbf{m}) | \mathbf{y}_s \odot \mathbf{m})
\end{split}
\end{equation}
where $\mathcal{M}$ is the set of possible binary masks. 
Through Eqn. \eqref{eqn:bert_gpt}, modeling the conditional structural constraint $p_s(\mathbf{y}^{\setminus k}_s |y^k_s)$
can be equivalently interpreted as learning the condition of \textit{masked pixels} on the given \textit{unmask pixel}.
To increase the modeling capability of the conditional structure network, three different strategies of the binary mask are adopted during training.
First, the binary mask only contains one unmasked pixel to model the condition structural constraint $p_s(\mathbf{y}^{\setminus k}_s |y^k_s)$.
Second, the binary mask does not contain any unmasked pixels (a zero mask). In this case, the model is going to learn the likelihood of the segmentation map $p_s(\mathbf{y}_s)$.
Third, the binary mask contains more than one unmasked pixel that aims to increase the generalizability of the conditional structure network in modeling segmentation structures conditioned on the unmasked pixels.

To model conditional structure network $G$ in a parallel fashion, the network $G$ is designed as a Transformer.
In particular, considering each pixel as a token, the network $G$ is formed as the Transformer %
with $L$ self-attention blocks where each block is designed in a residual style and the layer norms are applied to both the multi-head self-attention and multi-perceptron layers.
By this design, the spatial relationship and structural dependencies 
can be modeled by the self-attention mechanism.
To effectively optimize the network $G$, we adopt the learning tactic of Image-GPT \cite{chen2020generative}.

\vspace{-2mm}
\section{Experiments}
In this section, we present our experimental results on two standard benchmarks, i.e., SYNTHIA $\to$ Cityscapes and GTA5 $\to$ Cityscapes. First, we review datasets and our implementation, followed by analyzing the effectiveness of our approach to fairness improvement in ablation studies. Finally, we compare our experimental results with prior SOTA domain adaptation approaches. The performance of segmentation models is evaluated using the mean Intersection over Union (mIoU) and the IoU's standard deviation. 

\subsection{Datasets and Implementation}
\noindent
\textbf{Cityscapes} \cite{cordts2016cityscapes}, a real-world dataset collected in European, 
consists of  $3,975$ urban images with high-quality, dense annotations of $30$ categories. 
The license of Cityscapes is available for academic and non-commercial purposes.

\noindent
\textbf{SYNTHIA} \cite{Ros_2016_CVPR} is a synthetic dataset for the semantic segmentation task generated from a virtual world. There are $9,400$ pixel-level labeled RGB images in SYNTHIA with 16 standard classes overlapping with Cityscapes. 
The license of SYNTHIA was registered under Creative Commons Attribution-NonCommercial-ShareAlike 3.0. 

\noindent
\textbf{GTA5} \cite{Richter_2016_ECCV}, a synthetic dataset generated from the game engine, contains $24,966$ high-resolution, densely labeled images created for the semantic segmentation task. There are $19$ standard classes between GTA5 and Cityscapes. The GTA5 dataset is protected under the MIT License.

\noindent \textbf{Implementation} Two different segmentation architectures are used in our experiments, i.e., (1) DeepLab-V2 \cite{chen2018deeplab} with the Resnet-101 backbone and (2) Transformer with the MiT-B3 backbone \cite{xie2021segformer}. The Transformer design of \cite{chen2020generative} has been adapted to our conditional network structure $G$. Our framework is implemented in PyTorch and trained on four 48GB-VRAM NVIDIA Quadro P8000 GPUs. The model is optimized by the SGD optimizer \cite{Bottou10large-scalemachine} with learning rate $2.5 \times 10^{-4}$, momentum $0.9$, weight decay $10^{-4}$, and batch size of $4$ per GPU. The image size is set to $1280 \times 720$ pixels. In the proposed FREDOM framework, the learning strategies and sampling techniques of \cite{daformer, Araslanov:2021:DASAC} are adopted for the self-supervised loss $\mathcal{L}_t$ to train our model. Our implementation is further detailed in the supplementary.

\vspace{-2mm}
\subsection{Ablation Study}

\begin{table*}[!ht]
    \centering
    \caption{Effectiveness of our FREDOM (DeepLab-V2) approach to fairness improvement. There are three configurations:
    (A) Model without $\mathcal{L}_{Class}$ and $\mathcal{L}_{Cond}$. 
    (B) \textbf{Fairness Model} with $\mathcal{L}_{Class}$ only. 
    (C) \textbf{Fairness Model} with $\mathcal{L}_{Class}$ and $\mathcal{L}_{Cond}$. 
    }
    \label{tab:ablation}
    \vspace{-3mm}
    \setlength{\tabcolsep}{2.5pt}
    \resizebox{\textwidth}{!}{%
    \begin{tabular}{c c | c c c c c c | c c c c c c c c c c c c c | c c}
\hline

\multicolumn{2}{c|}{\multirow{2}{*}{Configuration}}  & \multicolumn{6}{c|}{Majority Group} & \multicolumn{13}{c|}{Minority Group} & \multirow{2}{*}{mIoU}          & \multirow{2}{*}{STD}  \\
 \cline{3-21}
   & & Road          & Build.        & Veget.        & Car           & S.Walk        & Sky           & Pole          & Person        & Terrain       & Fence         & Wall          & Sign          & Bike          & Truck         & Bus           & Train         & Tr.Light      & Rider         & M.bike        &           &          \\
\toprule                              
\multicolumn{23}{c}{SYNTHIA $\to$ Cityscapes}        \\
\toprule

\multirow{3}{*}{\begin{tabular}{@{}c@{}} Without \\ Adaptation \end{tabular} }
    & (A)          & 64.9          & 71.5          & 73.1          & 62.9          & 26.1          & 71.0          & 21.7          & 48.4          & $-$           & 0.2           & 3.0           & 0.2           & 35.6          & $-$           & 27.9          & $-$           & 0.1           & 20.7          & 12.0          & 33.7          & 27.8          \\
    & (B)          & 65.0          & 72.1          & 64.9          & 65.8          & 31.9          & 66.6          & 23.2          & 49.6          & $-$           & 0.2           & 5.0           & 2.5           & 31.7          & $-$           & 26.8          & $-$           & 2.4           & 21.3          & 18.7          & 34.4          & 26.1          \\
    & (C)          & 65.2          & 73.3          & 65.4          & 69.0          & 32.2          & 67.7          & 34.5          & 50.0          & $-$           & 0.3           & 17.5          & 3.5           & 39.9          & $-$           & 27.0          & $-$           & 3.9           & 21.9          & 18.5          & 36.7          & 25.4          \\
\cline{3-23}

\multirow{3}{*}{\begin{tabular}{@{}c@{}} With \\ Adaptation \end{tabular} }
    & (A)          & 84.9          & 85.7          & 86.4          & 86.8          & 44.9          & 88.6          & 45.8          & 69.3          & $-$           & 2.5           & 31.0          & 40.5          & 57.1          & $-$           & 45.9          & $-$           & 48.9          & 31.4          & 47.4          & 56.1          & 25.3          \\
    & (B)          & 84.8          & 85.8          & 86.4          & 86.8          & 45.2          & 88.9          & 47.6          & 70.1          & $-$           & 2.6           & 31.3          & 43.0          & 58.5          & $-$           & 46.0          & $-$           & 51.9          & 34.1          & 49.2          & 57.0          & 24.9          \\
    & \textbf{(C)} & \textbf{86.0} & \textbf{87.0} & \textbf{87.1} & \textbf{87.1} & \textbf{46.3} & \textbf{89.1} & \textbf{48.7} & \textbf{71.2} & \textbf{$-$}  & \textbf{5.3}  & \textbf{33.3} & \textbf{46.8} & \textbf{59.9} & \textbf{$-$}  & \textbf{54.6} & \textbf{$-$}  & \textbf{53.4} & \textbf{38.1} & \textbf{51.3} & \textbf{59.1} & \textbf{24.0} \\                                                    
\toprule                              
\multicolumn{23}{c}{GTA5 $\to$ Cityscapes}        \\
\toprule

 \multirow{3}{*}{\begin{tabular}{@{}c@{}} Without \\ Adaptation \end{tabular} }
 & (A)          & 75.8          & 77.2          & 81.3          & 49.9          & 16.8          & 70.3          & 25.5          & 53.8          & 24.6          & 21.0          & 12.5          & 20.1          & 36.0          & 17.2          & 25.9          & 6.5           & 30.1          & 26.4          & 25.3          & 36.6          & 24.0          \\
 & (B)          & 76.2          & 77.7          & 83.0          & 51.2          & 17.5          & 71.5          & 26.0          & 52.5          & 28.5          & 21.7          & 13.7          & 22.6          & 37.7          & 18.4          & 26.5          & 7.1           & 40.7          & 27.1          & 26.3          & 38.2          & 23.6          \\
 & (C)          & 77.1          & 79.4          & 84.7          & 52.9          & 18.5          & 72.3          & 28.6          & 54.4          & 33.8          & 22.5          & 15.6          & 23.7          & 38.9          & 19.7          & 27.1          & 7.9           & 41.6          & 28.6          & 28.0          & 39.7          & 23.6          \\

\cline{3-23}
\multirow{3}{*}{\begin{tabular}{@{}c@{}} With \\ Adaptation \end{tabular} }
 & (A)          & 90.3          & 87.2          & 88.1          & 88.6          & 53.5          & 87.3          & 44.4          & 67.3          & 42.2          & 28.5          & 41.1          & 50.1          & 54.4          & 52.5          & 56.9          & 33.7          & 48.9          & 33.1          & 42.6          & 57.4          & 20.9          \\
 & (B)          & 90.6          & 87.3          & 88.1          & 88.8          & 53.7          & 87.4          & 44.9          & 67.7          & 42.3          & 28.6          & 41.9          & 52.9          & 57.6          & 55.2          & 57.5          & 47.6          & 50.8          & 36.9          & 44.9          & 59.2          & 19.8          \\
 & \textbf{(C)} & \textbf{90.9} & \textbf{87.8} & \textbf{88.6} & \textbf{89.7} & \textbf{54.1} & \textbf{89.5} & \textbf{45.2} & \textbf{68.8} & \textbf{42.6} & \textbf{32.6} & \textbf{44.1} & \textbf{57.1} & \textbf{58.1} & \textbf{58.4} & \textbf{62.6} & \textbf{55.3} & \textbf{51.4} & \textbf{40.0} & \textbf{47.7} & \textbf{61.3} & \textbf{19.1} \\
 
\toprule
\end{tabular}
    }
\vspace{-6mm}
\end{table*}

\begin{figure}[t]
    \centering
    \includegraphics[width=0.45\textwidth]{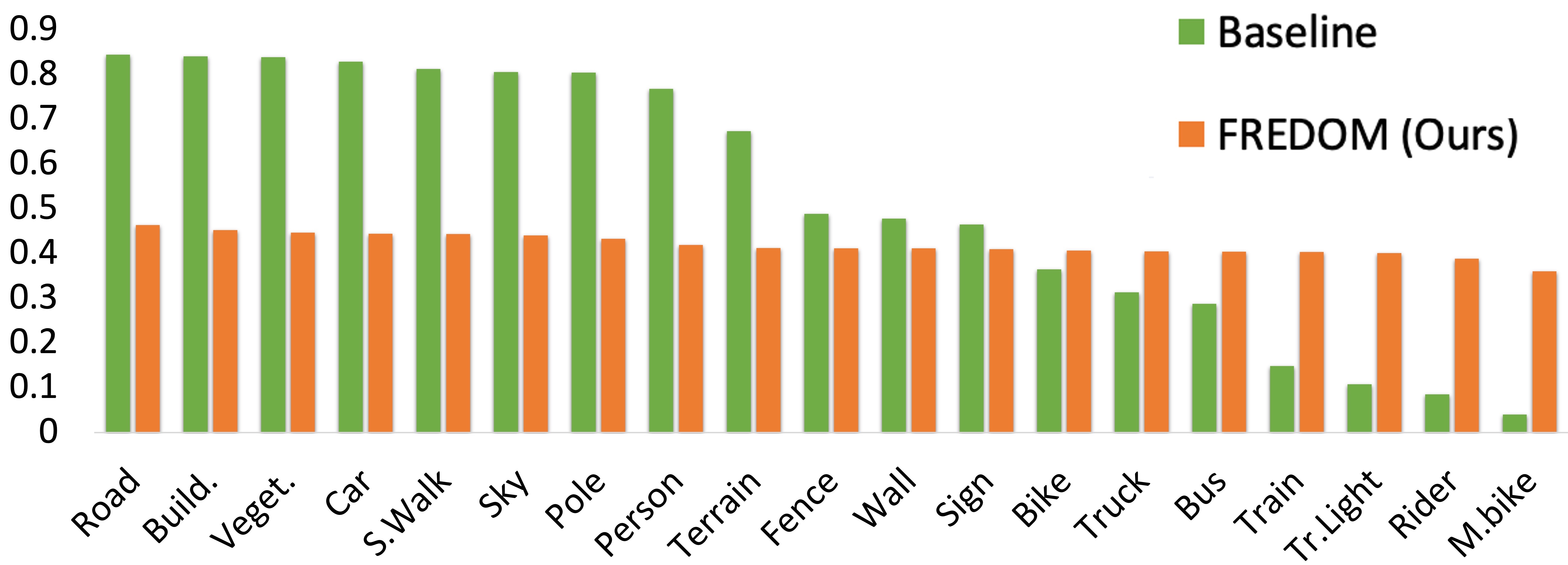}
    \vspace{-3.8mm}
    \caption{\textbf{The Mean Magnitude of Normalized Gradients Updated for Each Class.} Configuration (A) is used as the baseline.}
    \vspace{-6mm}
    \label{fig:grad_per_class}
\end{figure}

Our ablation studies evaluate DeepLab-V2 models on two benchmarks under two settings, i.e., With and Without Adaptation.
Each setting has three configs, i.e., (A) Model without $\mathcal{L}_{Class}$ and $\mathcal{L}_{Cond}$, (B) \textbf{\textit{Fairness model}} with only $\mathcal{L}_{Class}$, and (C) \textbf{\textit{Fairness model}} with $\mathcal{L}_{Class}$ and $\mathcal{L}_{Cond}$. 

\noindent \textbf{Does Adaptation Improve the Fairness?} 
We evaluate the impact of Domain Adaptation in improving the fairness of classes in the minor group. As shown in Tab. \ref{tab:ablation}, domain adaptation significantly improves fairness. In particular, without adaptation, the segmentation models trained only on the source data retain low performance in classes in the minor group, i.e., Traffic Light, Sign, and Fence. However, with our fairness domain adaptation approach, the overall accuracy and individual IoU of classes in the minor group are significantly boosted. In particular, the mIoU accuracy of segmentation models has been improved by $+22.4\%$ and $+21.6\%$ on SYNTHIA $\to$ Cityscapes and GTA5 $\to$ Cityscapes benchmarks. The model's fairness has been improved. Meanwhile, the IoU's STD of classes has been reduced by $1.4\%$ and $4.5\%$ on two benchmarks, respectively.

\noindent \textbf{Does Class Distributions Matter to Fairness Improvement?}
As shown in Table \ref{tab:ablation}, the fairness treatment from the class distribution loss $\mathcal{L}_{Class}$ contributes a significant improvement to both the overall performance and accuracy of classes in the minority group. In particular,  the IoU accuracy of each class in configuration (B) is improved compared to the one in configuration (A) in both with and without adaptation settings. Specifically, in the adaptation setting on benchmark SYNTHIA $\to$ Cityscapes, the class distribution loss $\mathcal{L}_{Class}$ has boosted the performance of classes in the minority group, e.g., Traffic Light (from $48.9\%$ to $51.9\%$),  Sign (from $40.5$ to $43.0\%$), Pole (from $45.8\%$ to $48.6\%$).
Without adaptation, improvement is also observed. Moreover, the standard deviation of IoU over classes has been reduced. It shows that the model's fairness has been promoted. Similarly, the performance of models on benchmark GTA5 $\to$ Cityscapes is also consistently improved. 

\noindent
\textbf{Does the Conditional Structure Constraint Contribute to Fairness Improvement?}
Configuration (C) in Table \ref{tab:ablation} reports experimental results of our model using conditional structure constraint loss $\mathcal{L}_{Cond}$. Results in Table \ref{tab:ablation} have shown the de facto role of the conditional structure constraint in performance improvement. Indeed, it enhances the IoU accuracy of each class in the minority group. For example, the average IoU accuracy of Fences, Pole, Traffic Light, and Sign has been improved by $2.3\%$. Overall, the performance of segmentation models has been improved by a notable margin, i.e., $+2.1\%$ and $2.7\%$ on SYNTHIA $\to$ Cityscapes and GTA5 $\to$ Cityscapes, respectively. The difference in performance between classes is reduced, illustrated by the decrease of the IoU's standard deviation, which means the model's fairness is improved notably.

\noindent \textbf{Does the Network Design Improve the Fairness?} Table \ref{tab:sota} illustrates the results of our approach using DeepLab-V2 and Transformer networks. As in our results, the performance of segmentation models using a more powerful backbone, i.e., Transformer, outperforms the models using DeepLab-V2. The performance of classes in the minority group has been improved notably, e.g., the performance of classes Fence, Traffic Light,  Sign, and Pole has been improved to $9.3\%$, $65.1\%$, $60.1\%$, and $57.3\%$ on the SYNTHIA $\to$ Cityscapes benchmark. The major improvements in the performance of overall and individual classes are also perceived in the GTA5 $\to$ Cityscapes benchmark. Also, the standard deviation of IoU over classes has been majorly reduced by $3.3\%$, illustrating that fairness has been promoted.

\noindent \textbf{Does the Model Fairly Treat all Class During Training?} Fig. \ref{fig:grad_per_class} visualizes the gradients produced w.r.t each class in the domain adaptation setting. In particular, we take a subset in Cityscapes and compute the normalized gradients updated for each class. The model with our proposed approach tends to update gradients for each class fairly. Meanwhile, without using our fairness method, the gradients of classes in the minority group are dominated by the ones in the majority group, which could result in models' unfair behaviors.

\subsection{Comparison with SOTA Approaches}

 \begin{table*}
        \centering
        \caption{Comparison of Semantic Segmentation Performance with UDA Methods Using DeepLab-V2 (\textbf{DL-V2}) and Transformer (\textbf{Trans.}).}
        \label{tab:sota}
        \setlength{\tabcolsep}{2.5pt}
        \vspace{-3mm}
        \resizebox{\textwidth}{!}{%
        \begin{tabular}{l c|cccccc|ccccccccccccc|cc}
        \hline
         \multicolumn{2}{c|}{\multirow{2}{*}{Approach $\quad$ Network}}   & \multicolumn{6}{c|}{Majority Group} & \multicolumn{13}{c|}{Minority Group} & \multirow{2}{*}{mIoU}          & \multirow{2}{*}{STD}  \\
         \cline{3-21}
        & &  Road          & Build.        & Veget.        & Car           & S.Walk        & Sky           & Pole          & Person        & Terrain       & Fence         & Wall          & Sign          & Bike          & Truck         & Bus           & Train         & Tr.Light      & Rider         & M.bike        &          &    \\
         
        \toprule
        \multicolumn{23}{c}{SYNTHIA $\to$ Cityscapes} \\
        \toprule
        
    IntraDA \cite{pan2020unsupervised}&  DL-V2  & 84.3 & 79.5   & 80.0   & 78.0 & 37.7   & 84.1 & 24.9 & 57.2   & $-$     & 0.4   & 5.3  & 8.4  & 36.5 & $-$   & 38.1 & $-$   & 9.2      & 23.0  & 20.3   & 41.7 & 31.0 \\
    BiMaL \cite{truong2021bimal} &   DL-V2 & \textbf{92.8} & 81.5   & 82.4   & 85.7 & 51.5   & 84.6 & 30.4 & 55.9   & $-$     & 1.0   & 10.2 & 15.9 & 38.8 & $-$   & 44.5 & $-$   & 17.6     & 22.3  & 24.6   & 46.2 & 30.9 \\
    SAC \cite{Araslanov:2021:DASAC} &   DL-V2   & 89.3 & 85.6   & 87.1   & 87.0 & \textbf{47.3}   & 89.1 & 43.1 & 63.7   & $-$     & 1.3   & 26.6 & 32.0 & 52.8 & $-$   & 35.6 & $-$   & 45.6     & 25.3  & 30.3   & 52.6 & 27.9 \\
    
    ProDA \cite{zhang2021prototypical} &   DL-V2  & 87.8 & \textbf{84.6}   & \textbf{88.1}   & 88.2 & 45.7   &  84.4    & 44.0 & \textbf{74.2}   & $-$     & 0.6   & \textbf{37.1} & 37.0 & 45.6 & $-$   & 51.1 & $-$   & \textbf{54.6}     & 24.3  & 40.5   & 55.5 & 26.4 \\
    
    \textbf{FREDOM} & DL-V2 & 86.0 & \textbf{87.0}   & 87.1   & 87.1 & 46.3   & \textbf{89.1} & \textbf{48.7} & 71.2   & $-$     & \textbf{5.3}   & 33.3 & \textbf{46.8} & \textbf{59.9} & $-$   & \textbf{54.6} & $-$   & 53.4     & \textbf{38.1}  & \textbf{51.3}   & \textbf{59.1} & \textbf{24.0} \\
    
    \hline
    TransDA \cite{transda} & Trans.  & \textbf{90.4} & 86.4   & \textbf{90.3}   & \textbf{92.3} & \textbf{54.8}   & 93.0 & 53.8 & 71.2   & $-$     & 1.7   & 31.1 & 37.1 & 49.8 & $-$   & 66.0 & $-$   & 61.1     & 25.3  & 44.4   & 59.3 & 27.3 \\
    ProCST \cite{procst} & Trans.   & 84.3 & 87.7   & 86.1   & 87.6 & 41.1   & 87.9 & 50.7 & 74.7   & $-$     & 6.1   & 42.6 & 54.2 & 62.5 & $-$   & 61.4 & $-$   & 55.5     & 47.2  & 53.3   & 61.4 & 22.6 \\
    DAFormer  \cite{daformer}& Trans. & 84.5 & 88.4   & 86.0   & 87.2 & 40.7   & 89.8 & 50.0 & 73.2   & $-$     & 6.5   & 41.5 & 54.6 & 61.7 & $-$   & 53.2 & $-$   & 55.0     & 48.2  & 53.9   & 60.9 & 22.8 \\
    
    \textbf{FREDOM} & Trans.  & 89.4 & \textbf{89.3}   & 89.9   & 90.5 & 50.8   & \textbf{93.7} & \textbf{57.3} & \textbf{79.4}   & $-$     & \textbf{9.3}   & \textbf{48.8} & \textbf{60.1} & \textbf{68.1} & $-$   & \textbf{66.0} & $-$   & \textbf{65.1}     & \textbf{51.6}  & \textbf{62.3}   & \textbf{67.0} & \textbf{22.0} \\

        \toprule
        \multicolumn{23}{c}{GTA5 $\to$ Cityscapes}\\
        \toprule
        
    IntraDA \cite{pan2020unsupervised} & DL-V2  & 90.6 & 82.6   & 85.2   & 86.4 & 36.1   & 80.2 & 27.6 & 59.3   & 39.3    & 21.3  & 29.5 & 23.1 & 37.6 & 33.6  & 53.9 & 0.0   & 31.4     & 29.4  & 32.7   & 46.3 & 26.7 \\
    BiMaL \cite{truong2021bimal} &  DL-V2   & \textbf{91.2} & 82.7   & 85.4   & 86.6 & 39.6   & 80.8 & 29.6 & 59.7   & 44.0    & 25.2  & 29.4 & 25.5 & 36.8 & 38.5  & 47.6 & 1.2   & 34.3     & 30.4  & 34.0   & 47.3 & 25.9 \\
    SAC \cite{Araslanov:2021:DASAC} &   DL-V2    & 90.3 & 86.6   & 87.5   & 88.5 & 53.9   & 86.0 & 45.1 & 67.6   & 40.2    & 27.4  & 42.5 & 42.9 & 45.1 & 49.0  & 54.6 & 9.8   & 48.6     & 29.7  & 26.6   & 53.8 & 24.2 \\
    ProDA \cite{zhang2021prototypical} &  DL-V2  & 87.8 & 79.7   & 88.6   & 88.8 & \textbf{56.0}   & 82.1 & \textbf{45.6} & \textbf{70.7}   & \textbf{45.2}    & \textbf{44.8}  & \textbf{46.3} & \textbf{53.5} & 56.4 & 45.5  & 59.4 & 1.0   & 53.5     & 39.2  & 48.9   & 57.5 & 21.7 \\
    
    \textbf{FREDOM} & DL-V2 & 90.9 & \textbf{87.8}   & \textbf{88.6}   & \textbf{89.7} & 54.1   & \textbf{89.5} & 45.2 & 68.8   & 42.6    & 32.6  & 44.1 & \textbf{57.1} & \textbf{58.1} & \textbf{58.4}  & \textbf{62.6} & \textbf{55.3}  & 51.4     & \textbf{40.0}  & \textbf{47.7}   & \textbf{61.3} & \textbf{19.1} \\
    
    \hline
    TransDA \cite{transda} & Trans. & 94.7 & 89.2   & 90.4   & 92.5 & 64.2   & 93.7 & 50.1 & 76.7   & 50.2    & 45.8  & 48.1 & 40.8 & 55.4 & 56.8  & 60.1 & 47.6  & 60.2     & 47.6  & 49.6   & 63.9 & 19.1 \\
    ProCST \cite{procst} &  Trans.  & 95.8 & 89.8   & 90.2   & 92.3 & 69.6   & 93.0 & 49.8 & 72.2   & 50.3    & 45.0  & 55.8 & 63.3 & 63.1 & 72.2  & 78.8 & 65.1  & 56.8     & 44.9  & 56.4   & 68.7 & 17.1 \\
    DAFormer \cite{daformer} & Trans. & 95.7 & 89.4   & 89.9   & 92.3 & 70.2   & 92.5 & 49.6 & 72.2   & 47.9    & 48.1  & 53.5 & 59.4 & 61.8 & 74.5  & 78.2 & 65.1  & 55.8     & 44.7  & 55.9   & 68.3 & 17.3 \\
    \textbf{FREDOM} & Trans. & \textbf{96.7} & \textbf{90.9}   & \textbf{91.6}   & \textbf{94.1} & \textbf{74.8}   & \textbf{94.4} & \textbf{57.5} & \textbf{78.4}   & \textbf{52.1}    & \textbf{49.0}  & \textbf{58.1} & \textbf{71.4} & \textbf{68.9} & \textbf{83.9}  & \textbf{85.2} & \textbf{72.5}  & \textbf{63.4}     & \textbf{53.1}  & \textbf{62.8}   & \textbf{73.6} & \textbf{15.8} \\
        
        \bottomrule
        
        \end{tabular}
        }
        \vspace{-6mm}
\end{table*}

\noindent
\textbf{SYNTHIA $\to$ Cityscapes}
Table \ref{tab:sota} presents our experimental results using  DeepLab-V2 and Transformer compared to prior SOTA approaches. Our proposed approach achieves SOTA performance and outperforms prior methods using the same network backbone. 
Specifically, the mIoU accuracy of our approach using Transformer  is $67.0\%$ and higher than DAFormer \cite{daformer} by $+6.1\%$.
Although the results of several individual classes are slightly lower than prior methods, overall, the mIoU accuracy and performance of individual classes in the minor group have been significantly promoted.
Analyzing the mIoU accuracy of classes in the minor group, 
our results have been significantly improved compared to the prior SOTA method (i.e., DAFormer \cite{daformer}). In particular, the performance of Rider, Fence, Pole, Traffic Light, and Sign classes has been improved by $4.1\%$, $+2.8\%$, $+7.3\%$, $+10.1\%$, and $+5.5\%$, respectively. In addition, the IoU accuracy of classes in the major group is also slightly enhanced. For example, the IoU accuracy of Building, Car, Sidewalk, and Sky has been improved to $87.8\%$, $89.7\%$, $54.1\%$, and $89.5\%$, respectively. 
It is vital to highlight that, to enhance the performance of classes in the minority group, the model does not sacrifice its ability to identify classes in the majority group. Instead, to promote the model's fairness, our approach enhances its ability to segment classes in the minor group to reduce the difference in performance between classes in minor and major groups.

\noindent \textbf{GTA5 $\to$ Cityscapes}
As shown in Table \ref{tab:sota}, on the same network backbone, our FREDOM approach performs better than previous SOTA methods. In particular, our approach using Transformer achieves the mIoU accuracy of $73.6\%$, which is the SOTA result; meanwhile, the result of the prior method \cite{daformer} is $68.3\%$. Noticeably, the performance results have been significantly enhanced in the classes of the minority group, e.g., in comparison with DAFormer \cite{daformer}, the IoU accuracy of Rider, Motorbike, Pole, Traffic Light, and Sign has been increased by $+8.4$, $+6.9\%$, $7.9\%$, $+7.6\%$, and $+12.0\%$. The performance accuracy has also improved in the majority group classes. For example, the accuracy of Building, Car, Sidewalk, and Sky is brought up to $90.9\%$, $94.1\%$, $74.8\%$, and $94.4\%$. Our FREDOM approach has strengthened the model's ability to segment classes in the minor group to lessen the performance gap between minor and major groups. In addition, the IoU's standard deviation over classes has been decreased compared to prior methods, which means that fairness has been promoted. 

\begin{figure}[!b]
    \centering
    \vspace{-6mm}
    \includegraphics[width=0.45\textwidth]{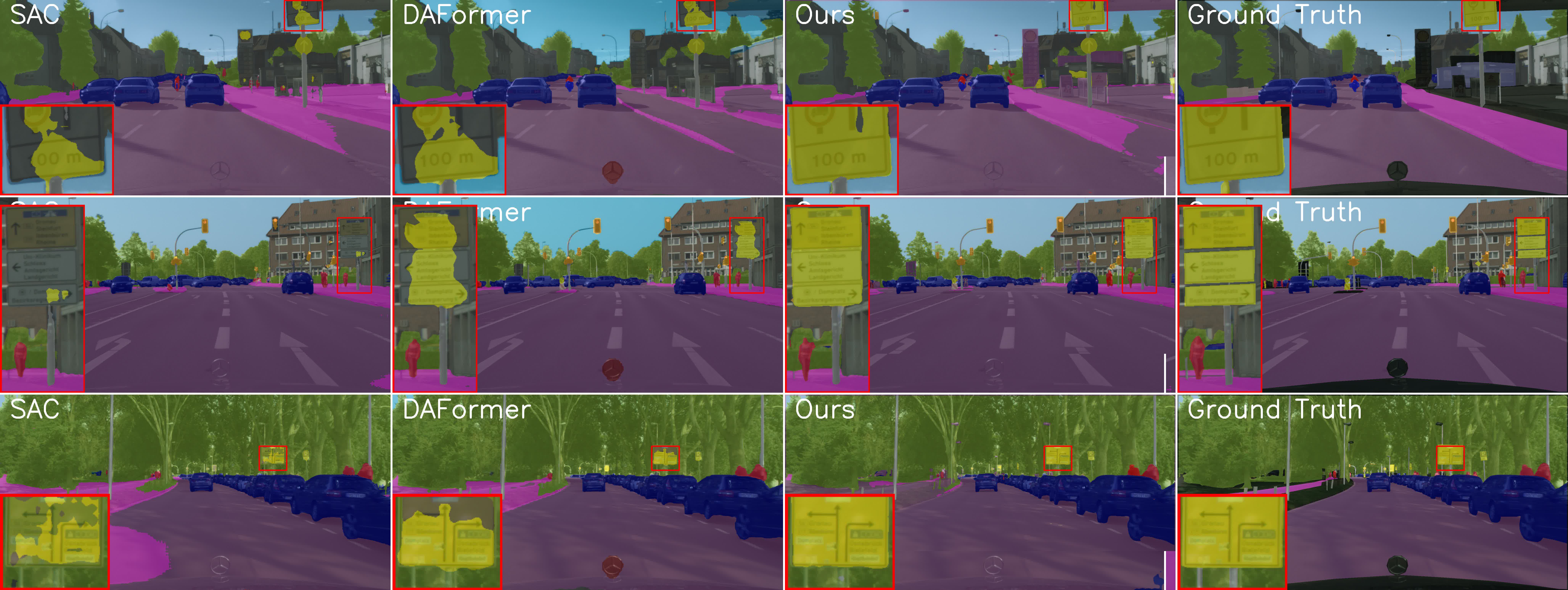}
    \vspace{-3mm}
    \caption{\textbf{Qualitative Results on SYNTHIA $\to$ Cityscapes}
    Columns 1-4 are the results of 
    SAC \cite{Araslanov:2021:DASAC}, and DAFormer \cite{daformer}, our FREDOM, and ground truths (Best view in $2\times$ zoom and color).
    }
    \label{fig:syn2city_result}
\end{figure}

\noindent \textbf{Qualitative Results}
Fig. \ref{fig:syn2city_result} illustrates our results of the SYNTHIA $\to$ Cityscapces experiment. Our approach produces better quality results than prior UDA methods. Particularly, a significant improvement can be observed from the predictions of classes in the minority group, e.g., the predicted segmentation of signs, persons, and poles is sharper. The model can well segment the classes in the minor group cogently and minimize the region of classes being erroneously classified. The borders between classes are accurately identified and predicted segmentation continuity has improved compared to prior works. Although our predictions contain some noise, the boundaries are still clear and correspond to the labels.
More comparisons of quantitative and qualitative results are available in the supplementary.

\vspace{-2mm}
\section{Conclusions and Limitations}
\vspace{-2mm}

This paper has presented the new fairness domain adaptation to semantic scene segmentation by analyzing the fairness treatment from class distributions. In particular, the conditional structural constraints have imposed the consistency of the predicted segmentation and modeled the structural information to improve the accuracy of segmentation models. Our ablation studies have analyzed different aspects affecting the fairness of segmentation models. It has also shown the effectiveness of our approach in terms of fairness improvement. Our FREDOM approach has achieved SOTA performance compared to prior methods.

\noindent \textbf{Limitations: }  One of the potential limitations in our approach is the computational cost of the conditional structural constraint $\mathcal{L}_{Cond}$. As the constraint is computed by conditional structure network $G$, it requires more computational resources and time during training. Also, our work only utilized specific self-supervised loss, network backbones, and hyper-parameters to support our hypothesis. However, different aspects of learning have yet to be fully exploited, e.g., learning hyper-parameters, additional unsupervised loss $\mathcal{L}_t$ (adversarial loss, self-supervised loss). These could be further exploited in our future work.

\small{
\noindent
\textbf{Acknowledgment} 
This work is supported by NSF Data Science, Data Analytics that are Robust and Trusted (DART), NSF WVAR-CRESH, and Googler Initiated Research Grant. We also acknowledge the Arkansas High Performance Computing Center for providing GPUs.
}
{\small
\bibliographystyle{ieee_fullname}
\bibliography{arxiv}
}

\end{document}